\documentclass[10pt,letterpaper,twocolumn]{article}
\usepackage[latin9]{inputenc}
\usepackage{color}
\usepackage{amsmath}
\usepackage{amssymb}
\usepackage{graphicx}
\usepackage[unicode=true,
 bookmarks=false,
 breaklinks=true,pdfborder={0 0 1},backref=section,colorlinks=false]
 {hyperref}
\hypersetup{
 pagebackref=true,letterpaper=true,colorlinks}

\makeatletter

\pdfpageheight\paperheight
\pdfpagewidth\paperwidth

\providecommand{\tabularnewline}{\\}

\usepackage{iccv}
\usepackage{times}
\usepackage{epsfig}

 \iccvfinalcopy % *** Uncomment this line for the final submission

\ificcvfinal\fi

\usepackage{algorithm}
\usepackage{algorithmic}

\makeatother

\begin{document}
%%%%%%%%% TITLE

\title{Semi-Latent GAN: Learning to generate and modify facial images from
attributes}

\author{Weidong Yin$^{*}$, Yanwei Fu$^{*\surd}$, Leonid Sigal$^{\dagger}$ and Xiangyang Xue$^{*}$ \\
$^{*}$ The school of Data Science, Fudan University, \\
$^{\dagger}$ Disney Research, \\
{\tt\small $\surd$ Corresponding authour: yanweifu@fudan.edu.cn}
}

\maketitle
\begin{abstract}
Generating and manipulating human facial images using high-level attributal
controls are important and interesting problems. The models proposed
in previous work can solve one of these two problems (generation or
manipulation), but not both coherently. This paper proposes a novel
model that learns how to both generate and modify the facial image
from high-level semantic attributes. Our key idea is to formulate
a Semi-Latent Facial Attribute Space (SL-FAS) to systematically learn
relationship between user-defined and latent attributes, as well as
between those attributes and RGB imagery. As part of this newly formulated
space, we propose a new model \textendash{} SL-GAN which is a specific
form of Generative Adversarial Network. Finally, we present an iterative
training algorithm for SL-GAN. The experiments on recent CelebA and
CASIA-WebFace datasets validate the effectiveness of our proposed
framework. We will also make data, pre-trained models and code available. 
\end{abstract}
\begin{figure}
\centering{}%
\begin{tabular}{c}
\includegraphics[scale=0.6]{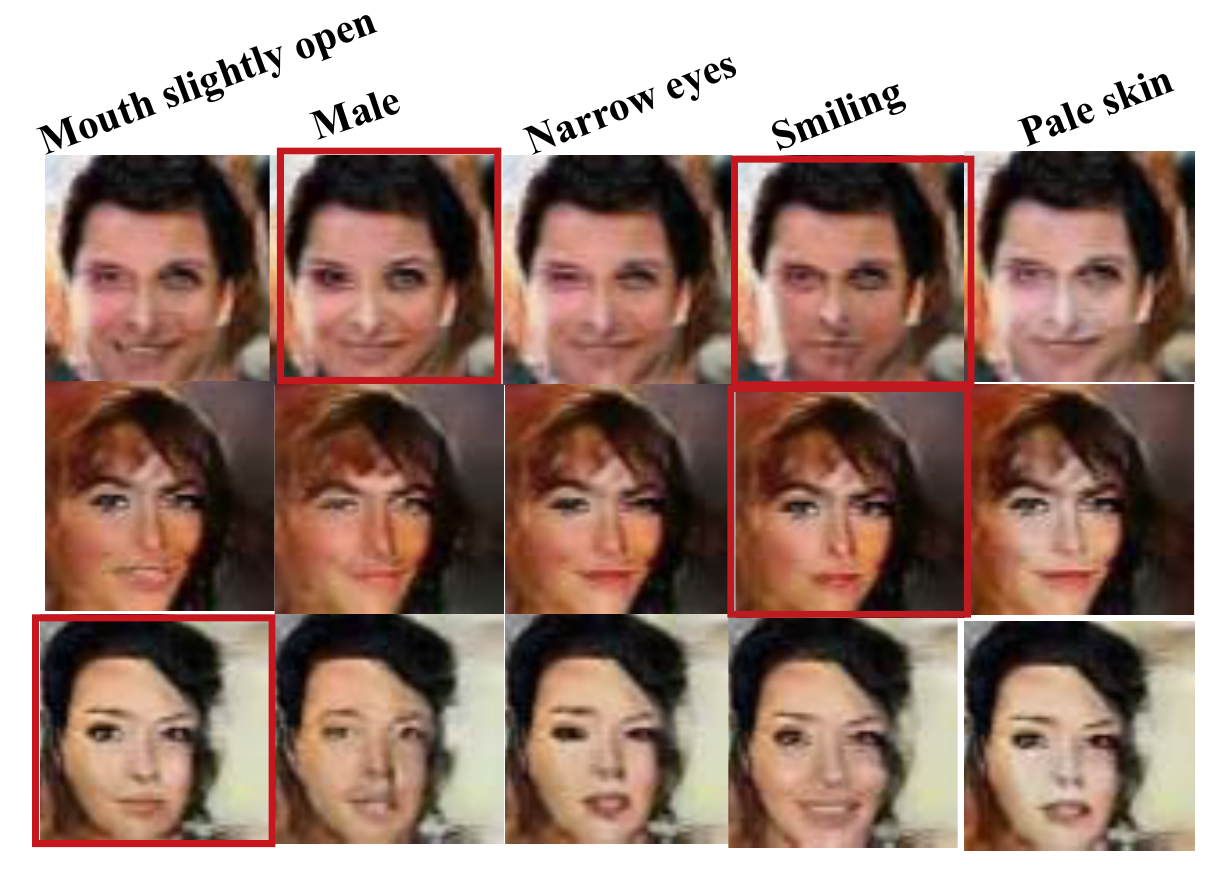}\tabularnewline
{\small{}{}(a) Generating facial images given an attribute vector}\tabularnewline
\includegraphics[scale=0.45]{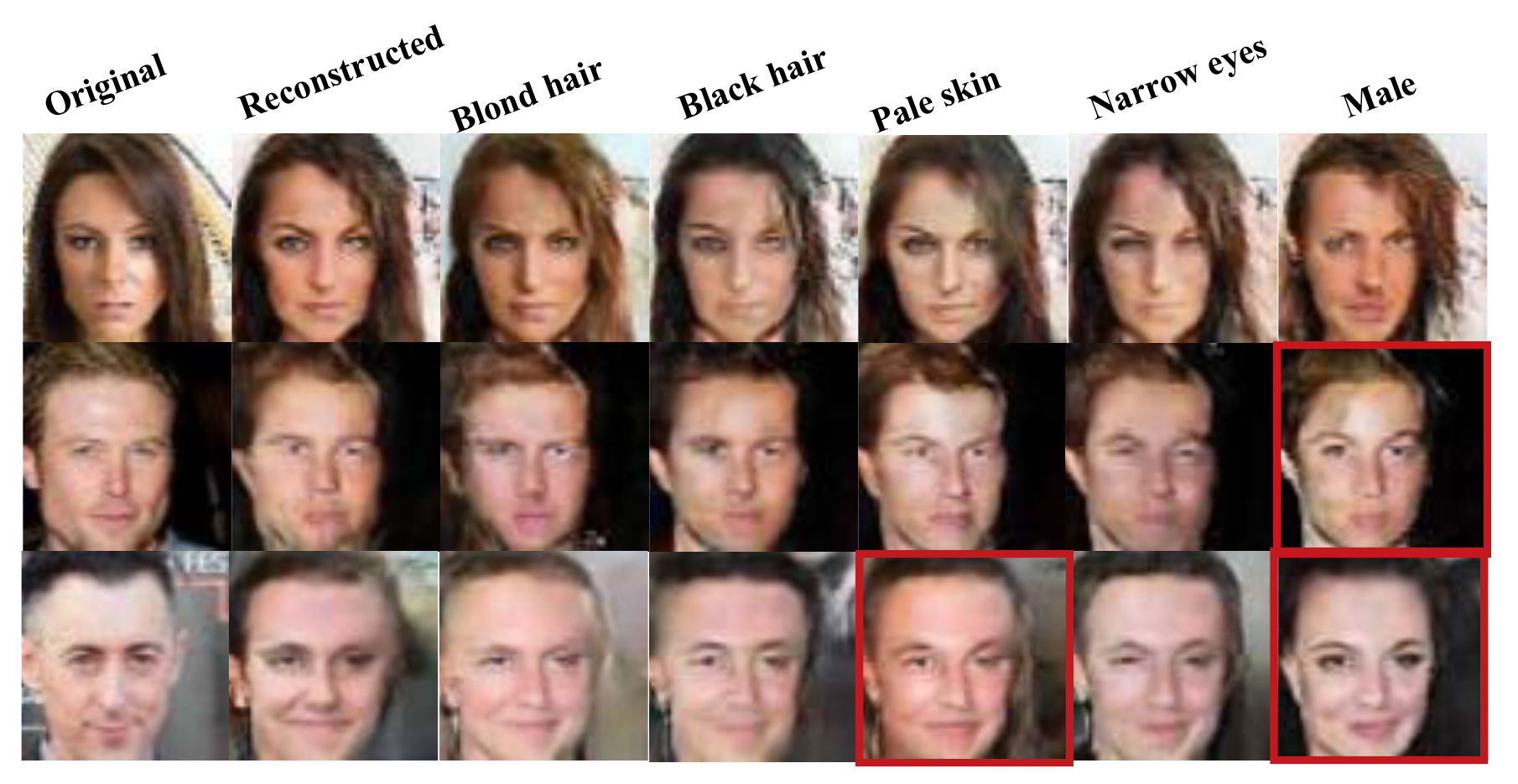}\tabularnewline
{\small{}{}(b) Modifying facial images by giving an attribute}\tabularnewline
\end{tabular}\caption{\label{fig:Facial-attribute-editing}\textbf{Examples of generation
and modification of facial images using attributes}. Red box indicates
that the inverse of an attribute is used in generation or modification
(\eg, in top row of (a) red boxes indicate {\em not male} and
{\em not smiling}). Note that in (a) we use an attribute vector
for generation, but only one attribute dimension is labeled in each
column for clarity of illustration. }
\vspace{-0.05in}
 
\end{figure}

\vspace{-0.07in}

\section{Introduction}

Analysis of faces is important for biometrics, non-verbal communication,
and affective computing. Approaches that perform face detection \cite{chen2014joint,li_face},
facial recognition \cite{ahonen2006face,schroff2015facenet,wen2016discriminative},
landmark estimation \cite{burgos2013robust,zhang2014facial}, face
verification \cite{sun2014deep,taigman2014deepface} and action coding
have received significant attention over the past 20+ years in computer
vision. However, an important problem of generation (or modification)
of facial images based on high-level intuitive descriptions remains
largely unexplored. For example, it would be highly desirable to generate
a realistic facial composite based on eyewitness' high level attributal
description (\eg, young, male, brown hair, pale skin) of the suspect.
Further, modifying facial attributes of a given person can help to
inform criminal investigation by visualizing how a suspect may change
certain aspects of their appearance to avoid capture. In more innocuous
use cases, modifying facial attributes may help a person visualize
what he or she may look like with a different hair color, style, makeup
and so on.

In this paper, we are interested in two related tasks: (i) {\em
generation} of facial images based on high-level attribute descriptions
and (ii) {\em modifying} facial images based on the high-level
attributes. The difference in the two tasks is important, for {\em
generation} one is interested in generating (a sample) image from
a distribution of facial images that contain user-specified attributes.
For modification, we are interested in obtaining an image of the pre-specified
subject with certain attributes changed. In both cases, one must take
care to ensure that the resulting image is of high visual quality;
in the {\em modification} case, however, there is an additional
constraint that identity of the person must be maintained. Intuitively,
solving these two tasks requires a generative model that models semantic
(attributal) space of faces and is able to decouple identity-specific
and identity-independent aspects of the generative process.

Inspired by \cite{yanweiPAMIlatentattrib}, we formulate the Semi-Latent
Facial Attribute Space (SL-FAS) which is a composition of two, user-defined
and latent, attribute subspaces. Each dimension of user-defined attribute
subspace corresponds to one human labeled interpretable attribute.
The latent attribute space is learned in a data-driven manner and
learns a compact hidden structure from facial images.

The two subspaces are coupled, making learning of SL-FAS challenging.
Recently, % based on Conditional Variational Auto-Encoder (CVAE), 
in \cite{yan2016eccv}, attribute-conditioned deep variational auto-encoder
framework was proposed that can learn latent factors (\emph{i.e}.
attributes) of data and generate images given those attributes. In
\cite{yan2016eccv}, only the latent factors are learned and the user-defined
attributes are given as input. Because of this, they can not model
the distribution of user-defined attributes for a given an image;
leading to inability to modify the image using semantic attributes.
Inspired by InfoGAN \cite{infogan}, we propose a network that jointly
models the subspace of user-defined and latent attributes.

In this paper, to jointly learn the SL-FAS, we propose a Semi-Latent
Generative Adversarial Network (SL-GAN) framework which is composed
of three main components, namely, (i) encoder-decoder network, (ii)
GAN and (iii) recognition network. In encoder-decoder network, the
encoder network projects the facial images into SL-FAS and the decoder
network reconstructs the images by decoding the attribute vector in
SL-FAS. Thus the decoder network can be used as a generator to generate
an image if given an attribute vector. The GAN performs the generator
and discriminator min-max game to ensure the generated images are
of good quality, by ensuring that generated images cannot be discriminated
from the real ones. Recognition network is the key recipe of jointly
learning user-defined and latent attributes from data. Particularly,
the recognition network is introduced to maximize the mutual information
between generated images and attributes in SL-FAS. Figure \ref{fig:Facial-attribute-editing}
gives the examples of generating and modifying the facial attributes.
As shown in the first and third rows of modification Figure \ref{fig:Facial-attribute-editing}
(b), our SL-GAN can modify the attributes of facial images in very
noise background. 

\vspace{0.07in}
\noindent \textbf{Contributions.} (1) To the best of our knowledge, there is
no previous work that can do both generation and modification of facial
images using visual attributes. Our framework only uses high-level
semantic attributes to modify the facial images. (2) Our SL-GAN can
systematically learn the user-defined and latent attributes from data
by formulating a semi-latent facial attribute space. (3) A novel recognition
network is proposed that is used to jointly learn the user-defined
and latent attributes from data. (4) Last but not the least, we propose
an iterative training algorithm to train SL-GAN in order to solve
two related and yet different tasks at the same time.

\section{Related Work}

\noindent \textbf{Attribute Learning.} Attribute-centric semantic
representations have been widely investigated in multi-task \cite{torralba2011app_share}
and transfer learning \cite{lampert13AwAPAMI}. Most early works \cite{lampert13AwAPAMI}
had assumed a space of user-defined namable properties as attributes.
User-defined facial attributes \cite{datta2011face_attrib,ehrlich2016facial,moon_attrb,wang2016walk,zhong2016face,kumar2009}
had also been explored. These attributes, however, is hard and prohibitively
expensive to specify, due to the manual effort required in defining
the the attributal set and annotating images with that set. To this
end, latent attributes \cite{yanweiPAMIlatentattrib} have been explored
for mining the attributes directly from data. It is important to note
that user-defined and latent attributes are complementary to one another
and can be used and learned simultaneously, forming a semi-latent
attribute space. Our SL-GAN model is a form of semi-latent attribute
space specifically defined for generation and modification of facial
images.

\vspace{0.07in}

\noindent \textbf{Deep Generative Image Modeling.} Algorithmic generation of
realistic images has been the focus of computer vision for some time.
Early attempts along these lines date back to 2006 with Deep Belief
Networks (DBNs) \cite{hinton2006science}. DBNs were successful for
small image patch generation, but failed to generalize to larger images
and more holistic structures. Recent models that address these challenges,
include auto-regressive models \cite{gregor2014icml,theis2015nips,oord2016nips},
Variational Auto-Encoders (VAEs) \cite{vaegan,cvae2016nips,yan2016eccv},
Generative Adversarial Networks (GANs) \cite{infogan,Denton2015nips,alexey2015cvpr,ishan2017iclr,gan2014,xun2017cvpr,kulkarni2015nips,anh2016play,scott2016nips,reed2016generative,reed2014icml},
and Deep Recurrent Attention Writer (DRAW) \cite{karol2015icml}.
InfoGAN \cite{infogan} and stackGAN \cite{xun2017cvpr} utilized
the recognition network to model the latent attributes, while our
SL-GAN extends the recognition network to jointly model user-defined
and latent attributes and thus our framework can both generate and
modify the facial images using attributes.

\vspace{0.07in}
\noindent \textbf{Semantic Image Generation.} More recently, there has been
a focus on generating images conditioned on semantic information (\eg,
text, pose or attributes). Reeds \emph{et al. } \cite{reed2016generative}
studied the problem of automatic synthesis of realistic images from
text using deep convolutional GANs. Yan \emph{et al.} \cite{yan2016eccv}
proposed an attribute-conditioned deep variational auto-encoder framework
that enables image generation from visual attributes. Mathieu \emph{et
al}. \cite{michael2016nips} learned the hidden factors within a set
of observations in the conditional generative model. However, their
frameworks can only generate the images rather than modifying an {\em
existing} image based on attributes or other form of semantic information.

\noindent \textbf{Image Editing and Synthesis}. Our work is also related to
previous work on image editing \cite{Kemelmacher2014cvpr,Guim2016nipsworkshop,wei2017arxiv,zhou2016eccv}.
The formulation in \cite{zhu2015eccv} also enables task of image
editing using GAN, where the user strokes are used to specify the
attributes being changed. In contrast, our SL-GAN does not need user
strokes as input. Very recently, two technical reports \cite{Guim2016nipsworkshop,wei2017arxiv}
can also enable modifying the attribute of facial images. \cite{wei2017arxiv}
proposed a GAN-based image transformation networks for face attribute
manipulation. In \cite{wei2017arxiv}, one trained model can only
modify one type attribute in facial images; in contrast, our SL-GAN
can learn to generate or modify many attributes simultaneously. In
fact, our models on CelebA dataset can generate or modify $17$ different
facial attributes all at ones.

\section{Semi-Latent GAN (SL-GAN)}

\subsection{Background}

\noindent 
\begin{figure}
\begin{centering}
\includegraphics[scale=0.14]{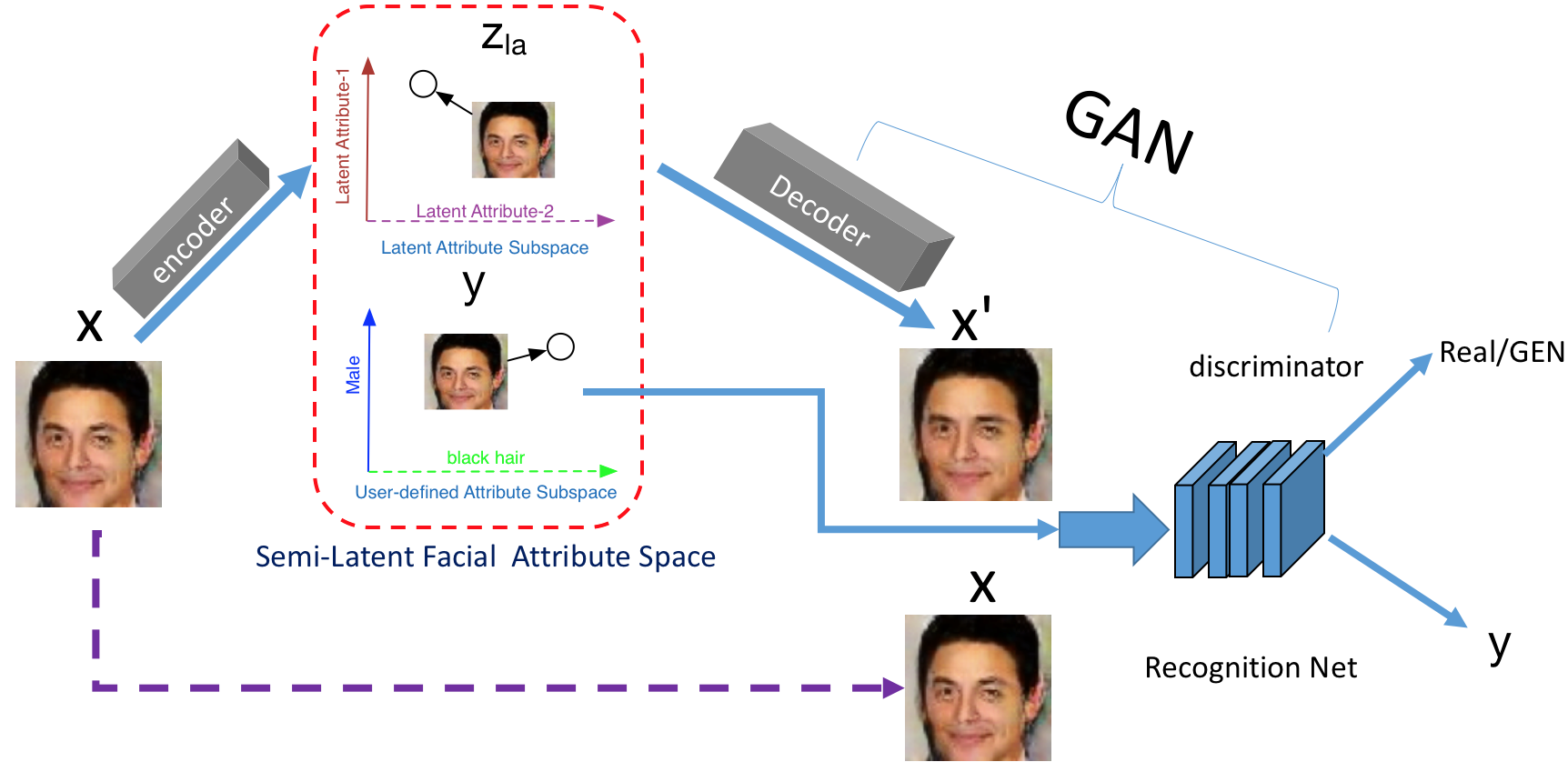} 
\par\end{centering}
\caption{\label{fig:Overview-of-our} \textbf{Overview of our SL-GAN.}}
\end{figure}

\noindent \textbf{GAN}~\cite{gan2014} aims to learn to discriminate
real data samples from generated samples; training the generator network
$G$ to fool the discriminator network $D$. GAN is optimized using
the following objective,

\begin{align}
\underset{G}{\mathrm{min}}\underset{D}{\mathrm{max}}\mathcal{L}_{GAN} & =\mathbb{E}_{x\sim p_{data}\left(x\right)}\left[\mathrm{log}D\left(x\right)\right]\label{eq:gan}\\
+ & \mathbb{E}_{z\sim p_{prior}\left(z\right)}\left[\mathrm{log}\left(1-D\left(G\left(z\right)\right)\right)\right],\nonumber 
\end{align}

\noindent where $p_{data}\left(x\right)$ is the distribution of real
data; $p_{prior}\left(z\right)$ is a zero-mean Gaussian distribution
$\mathcal{N}\left(\mathbf{0},\mathbf{1}\right)$. The parameters of
$G$ and $D$ are updated iteratively in training. The loss function
for generator and discriminator are $\mathcal{L}_{D}=\mathcal{L}_{GAN}$
and $\mathcal{L}_{G}=-\mathcal{L}_{GAN}$ respectively. To generate
an image, generator draws a sample $z\sim p_{prior}\left(z\right)=\mathcal{N}\left(\mathbf{0},\mathbf{1}\right)$
from prior ({\em a.k.a.}, noise distribution) and then transforms
that sample using a generator network $G$, \ie, $G(z)$.

\vspace{0.07in}
\noindent \textbf{InfoGAN} \cite{infogan} decomposes the input noise of GAN
into latent representation $y$ and incompressible noise $z$ by maximizing
the mutual information $I\left(y;G\left(z,y\right)\right)$ and it
ensures no loss information of latent representation during generation.
The mutual information term can be described as the recognition loss,

\noindent 
\[
\mathcal{L}_{rg}=-\mathbb{E}_{x\sim G\left(z,y\right)}\left[\mathbb{E}_{y\sim p_{data}\left(y\mid x\right)}\left[\log Q(y\mid x)\right]\right]
\]

\noindent where $Q\left(y\mid x\right)$ is an approximation of the
posterior $p_{data}\left(y\mid x\right)$. Thus the parameters of
the generator $G$ can thus be updated as $\mathcal{L}_{G_{InfoGAN}}=\mathcal{L}_{G}-\mathcal{L}_{rg}$.
InfoGAN can learn the disentangled, interpretable and meaningful representations
in a completely unsupervised manner.

\vspace{0.07in}
\noindent \textbf{VAEGAN}~\cite{vaegan} combines the VAE into GAN and replaces
the element-wise errors of GAN with feature-wise errors of VAEGAN
in the data space. Specifically, it encodes the data sample $x$ to
latent representation $z$: $z\sim\mathrm{Enc}\left(x\right)=p_{enc}\left(z\mid x\right)$
and decodes the $z$ back to data space: $\tilde{x}\sim\mathrm{Dec}\left(x\mid z\right)=p_{dec}\left(x\mid z\right)$.
We can define the loss functions of the regularization prior as $\mathcal{L}_{prior}=KL\left(q_{enc}\left(z\mid x\right)\parallel p_{prior}\left(z\right)\right)$
, $q_{enc}\left(z\mid x\right)$ is the approximation to the true
posterior $p_{dec}\left(z\mid x\right)$. The reconstruction error
is $\mathcal{L}_{recon}^{D_{l}}=-\mathbb{E}_{q_{enc}\left(z\mid x\right)}\left[\mathrm{log}p_{dec}\left(D_{l}\left(x\right)\mid z\right)\right]$
where $D_{l}\left(x\right)$ is hidden representation of $l-th$ layer
of the discriminator. Thus this loss minimizes the sum of the expected
log likelihood of the data representation of $l-th$ layer of discriminator.
Thus the loss function of VAEGAN is

\noindent 
\begin{equation}
\mathcal{L}_{VAEGAN}=\mathcal{L}_{GAN}+\mathcal{L}_{recon}^{D_{l}}+\mathcal{L}_{prior}\label{eq:vaegan}
\end{equation}

\noindent However the latent representation $z$ is totally unsupervised;
there is no way to explicitly control the attributes over the data
or modify facial images using attributes.

\vspace{0.07in}

\noindent \textbf{CVAE}~\cite{yan2016eccv,cvae2016nips} is the conditional
VAE. The independent attribute variable $y$ is introduced to control
the generating process of $x$ by sampling from $p\left(x\mid y,z\right)$;
where $p\left(y,z\right)=p\left(y\right)p\left(z\right)$. The encoder
and decoder networks of CVAE are thus $z\sim Enc\left(x\right)=q_{enc}\left(z\mid x\right)$
and $\text{\ensuremath{\tilde{x}}}\sim Dec\left(z,y\right)=p_{dec}\left(x\mid z,y\right)$.
The variable $y$ is introduced to control the generate process of
$x$ by sampling from $p\left(x\mid y,z\right)$; where $p\left(y,z\right)=p\left(y\right)p\left(z\right)$.
Nevertheless, $y$ is still sampled from data, rather than being directly
optimized and learned from the data as our SL-GAN. Thus $y$ can be
used to modify the attributes similar to proposed SL-GAN.

\subsection{Semi-Latent Facial Attribute Space}

The input noise of GAN can be further decomposed into two parts: (1)
User-defined~attributes~$y$ are the manually annotated attributes
of each image $x$, \emph{i.e}. $y\sim p_{data}\left(y\mid x\right)$;
(2) Latent~attributes~$z_{la}$ indicate the attributes that should
be mined from the data\footnote{Note that the latent attribute $z_{la}$ also includes the incompressible
noise, which is not explicitly modelled due to less impact to our
framework.} in a data-driven manner, \emph{i.e.}, $z_{la}\sim p_{data}\left(z_{la}\mid x\right)$.

Mathematically, the $y$ and $z_{la}$ can be either univariate or
multivariate; and these attributes are mutual independent, \emph{i.e.},
$p\left(y,z_{la}\right)=p\left(y\right)p\left(z_{la}\right)$. Each
dimension of $y$ is clipped with one type of facial attribute annotated
in real images; our SL-GAN train $y$ in a supervised manner. In contrast,
each dimension of $z_{la}$ is trained totally unsupervised. We define
semi-latent facial attribute space to combine both $y$ attributes
and the latent $z_{la}$ attributes.

With the decomposed input noise, the form of generator is $G\left(z_{la},y\right)$
now. Directly learning $z_{la}$ from input data will lead to the
trivial solution that the generator is inclined to ignore the latent
attributes as $p_{G}\left(x\mid z_{la},y\right)=p_{G}\left(x\mid y\right)$.
In contrast, we maximize the mutual information between the attributes
$z_{la}$ and the generator distribution $G\left(z_{la},y\right)$,
which can be simplified as minimizing the recognition loss for the
attribute $y$ and $z_{la}$.

It is important to jointly learn the attributes $y$ and $z_{la}$;
and make sure that $z_{la}$ can represent un-modeled aspects of the
input facial images rather than re-discovering $y$. The ``rediscovering''
means that some dimensions of $z_{la}$ have very similar distribution
as the distribution of $y$ over the input images, \ie, the same
patterns in $y$ repeatedly discovered from latent attributes $z_{la}$

\subsection{Semi-Latent GAN}

Our SL-GAN is illustrated in Fig. \ref{fig:Overview-of-our}; it is
composed of three parts, namely, encoder-decoding network, GAN and
recognition network. The user-defined and latent attributes are encoded
and decoded by the encoder-decoder network. Recognition network helps
learn the SL-FAS from data. In our SL-GAN, the recognition network
and discriminator shares the same network structure and have different
softmax layer at the last layer. The loss functions of the generator
$G_{SL-GAN}$ and discriminator $D_{SL-GAN}$ are thus, 
\begin{align}
\mathcal{L}_{G_{SL-GAN}} & =\mathcal{L}_{G}+\lambda_{1}\left(\mathcal{L}_{rg_{z}}+\mathcal{L}_{rg_{y}-G}\right)+\lambda_{2}\mathcal{L}_{recon}^{D_{l}}\label{eq:g_sl-gan}\\
\mathcal{L}_{D_{SL-GAN}} & =\mathcal{L}_{D}+\left(\mathcal{L}_{rg_{y}-D}+\mathcal{L}_{rg_{z}}\right)\label{eq:d_sl-gan}
\end{align}

\noindent where $\mathcal{L}_{rg_{z}}$ is the recognition loss on
$z_{la}$. For recognition loss on $y$, we use $\mathcal{L}_{rg_{y}-D}$
and $\mathcal{L}_{rg_{y}-G}$ as the loss for the discriminator and
generator respectively. We also define the decoder loss as $\mathcal{L}_{dec}=\mathcal{L}_{G_{SL-GAN}}$
; and the encoder loss as $\mathcal{L}_{enc}=\mathcal{L}_{VAE}$.

\vspace{0.07in}

\noindent \textbf{Encoder loss $\mathcal{L}_{enc}$} is the sum of
reconstruction error of the variational autoencoder and a prior regularization
term over the latent distribution $z_{la}$; thus it is defined as
$\mathcal{L}_{VAE}=\mathcal{L}_{prior}+\mathcal{L}_{recon}$; and
the $\mathcal{L}_{prior}=KL\left(q_{enc}\left(z_{la}\mid x\right)\parallel p_{prior}\left(z_{la}\right)\right)$
measures the KL-divergence between approximate posterior $q_{enc}\left(z_{la}\mid x\right)$
and the prior $p_{prior}\left(z_{la}\right)$. The reconstruction
loss $L_{recon}=-\mathbb{E}_{z\sim q_{enc}\left(z_{la}\mid x\right),y\sim p_{data}\left(y\mid x\right)}\left[\log{p_{dec}\left(x\mid z_{la},y\right)}\right]$
measures loss of reconstructing generated images by sampling the attributes
in SL-FAS. Here, $q_{enc}\left(z_{la}\mid x\right)$ is an approximation
of $p_{enc}\left(z_{la}\mid x\right)$ parameterized by a neural network,
\emph{e.g.}, the encoder.

\vspace{0.07in}

\noindent \textbf{The recognition loss of $z_{la}$} is trained on
both generated data and real data. It aims at predicting the values
of latent attributes. Suppose the latent $z_{la}\sim p_{enc}\left(z\mid x\right)$
is sampled from the distribution of encoder network. The update steps
of generator and discriminator use the same recognition loss on $z_{la}$
defined as, 
\begin{align}
\mathcal{L}_{rg_{z}} & =-\mathbb{E}_{x\sim p_{data}\left(x\right),z\sim p_{enc}\left(z_{la}\mid x\right)}\left[\log\left(Q\left(z\mid x\right)\right)\right]\label{eq:recognize_z}\\
- & \mathbb{E}_{x\sim p_{dec}\left(x\mid z,y\right),y\sim p_{data}\left(y\mid x\right),z\sim p_{enc}\left(z_{la}\mid x\right)}\left[\log\left(Q\left(z\mid x\right)\right)\right]\nonumber \\
- & \mathbb{E}_{x\sim p_{dec}\left(x\mid y,z\right),y\sim p_{data}\left(y\right),z\sim p_{prior}\left(z_{la}\right)}\left[\mathrm{log}\left(Q\left(z\mid x\right)\right)\right]\nonumber \\
- & \mathbb{E}_{x\sim p_{dec}\left(x\mid z,y\right),y\sim p_{data}\left(y\right),z\sim p_{enc}\left(z_{la}\mid x\right)}\left[\log\left(Q\left(z\mid x\right)\right)\right],\nonumber 
\end{align}
where $-\mathbb{E}_{x\sim p_{data}\left(x\right),z\sim p_{enc}\left(z_{la}\mid x\right)}\left[\log\left(Q\left(z\mid x\right)\right)\right]$
measures the loss of predicting errors on real data; and rest three
term are the loss functions on generated data.\textcolor{red}{{}
}$Q\left(\text{\ensuremath{\cdot}}\right)$ is still an approximation
of the corresponding posterior data distribution. $p_{enc}\left(z_{la}\mid x\right)$
is the distribution of $z_{la}$ given $x$ parameterized by the encoder
network; $p_{data}\left(y\right)$ is the data distribution of $y$
on real data; $p_{data}\left(y\mid x\right)$ is the data distribution
of $y$ given $x$ on the real data; $p_{data}\left(x\right)$ is
the data distribution of $x$ on the real data; $p_{prior}\left(z_{la}\right)$
is the prior distribution of $z$ and we use the Gaussian distribution
$\mathcal{N}\left(\mathbf{0},\mathbf{1}\right)$; $p_{dec}\left(x\mid z_{la},y\right)$
is the distribution of $x$ given $z_{la}$ and $y$, and the distribution
is parameterized by the decoder network; $p_{enc}\left(z_{la}\mid x\right)$
is the distribution of $z$ given $x$ and the distribution parameterized
by the encoder network;

\noindent \vspace{0.07in}

\noindent \textbf{The recognition loss of $y$} can be trained on
real data and generated data. In training stage of discriminator $D_{SL-GAN}$,
we have the ground-truth attribute annotations for user-defined $y$
without using the generated data. The reason is that the quality of
generated data relies on the training quality of generator $G_{SL-GAN}$,
which is also trained from the real data; thus we will observe a phenomena
of ``semantic drift'' if the data is generated by a not well trained
generator. To that end, only the manually labeled attribute $y$ can
be used for updating $D_{SL-GAN}$; and the loss of updating $D$
is 
\begin{equation}
\mathcal{L}_{rg_{y}-D}\!=\!-\mathbb{E}_{x\sim p_{data}(x),y\sim p_{data}\left(y\mid x\right)}\left[\log\left(Q\left(y\mid x\right)\right)\right]\label{eq:recog_loss_D}
\end{equation}

$G_{SL-GAN}$ is corresponding to the decoder-network, and thus only
generated data can be used to compute the loss of generator,

\begin{equation}
\begin{array}{cc}
\mathcal{L}_{rg_{y}-G}=\\
-\mathbb{E}_{x\sim p_{dec}\left(x\mid y,z\right),z\sim p_{enc}\left(z\mid x\right),y\sim p_{data}\left(y\right)}\left[\log\left(Q\left(y\mid x\right)\right)\right]\\
-\mathbb{E}_{x\sim p_{dec}(x\mid y,z),z\sim p_{enc}\left(z\mid x\right),y\sim p_{data}\left(y\mid x\right)}\left[\log\left(Q\left(y\mid x\right)\right)\right]\\
-\mathbb{E}_{x\sim p_{dec}\left(x\mid y,z\right),z\sim p_{prior}\left(z\right),y\sim p_{data}\left(y\mid x\right)}\left[\log\left(Q\left(y\mid x\right)\right)\right]
\end{array}\label{eq:recog_loss_G}
\end{equation}
Note that intrinsically in each term in Eq (\ref{eq:recognize_z})
and (\ref{eq:recog_loss_G}) could be weighted by a coefficient. Here
we omit these coefficients for the ease of notation.

\subsection{The training algorithms}

Our SL-GAN aims at solving generation and modification of facial images
via attributes. The key ingredients are to learn a disentangled representation
of data-driven attributes $z_{la}$ and user-defined attributes $y$
unified in our SL-GAN framework. Our SL-GAN is composed of three key
components and aims at solving two related and yet very different
problems \textendash{} generation and modification of facial images.
The conventional way of training GAN does not work in our problem.
Thus we propose a new training algorithms to train our SL-GAN. Specifically,
one iteration of our algorithm needs to finish the following three
stages,

\vspace{0.07in}
\noindent \textbf{Learning facial image reconstruction}. This stage mainly updates
the encoder-decoder network and learns to reconstruct the image given
corresponding user-defined attributes with the steps of, 
\begin{itemize}
\item Sampling a batch of images $x\sim p_{data}(x)$ and attributes $y\sim p_{data}(y|x)$,
$z_{la}\sim p_{enc}\left(z\mid x\right)$ 
\item Updating the SL-GAN by minimizing the encoder $\mathcal{L}_{enc}$,
Decoder $\mathcal{L}_{dec}$ and Discriminator with $\mathcal{L}_{D_{SL-GAN}}$
iteratively. 
\end{itemize}
\noindent \textbf{Learning to modify the facial image.} We sample
an image $x$ and the attribute $y$ from all data. This stage trains
the SL-GAN to be able to modify the image $x$ by supplying $y$.
Note that the image $x$ does not necessarily have the attribute $y$.
Another important question here is how to keep the same identity of
sampled images when modifying the facial image; two strategies are
employed here: first, $z_{la}$ is sampled from $p_{enc}\left(z\mid x\right)$
parameterized by the encoder network which is not updated in this
sub-step; second, our SL-GAN minimizing the $\mathcal{L}_{recon}^{D_{l}}$
essentially to guarantee the identity of the same person. Thus this
step encourages the generator to learn a disentangled representation
of $y$ and $z_{la}$. 
\begin{itemize}
\item Learning to modify the attributes: Sample a batch of images $x\sim p_{data}(x)$
and the attribute $y\sim p_{data}(y)$, $z_{la}\sim p_{enc}\left(z\mid x\right)$; 
\item Update the SL-GAN by minimizing the Decoder $\mathcal{L}_{dec}$ and
Discriminator with $\mathcal{L}_{D_{SL-GAN}}$ iteratively; 
\end{itemize}
\noindent \textbf{Learning to generate facial images.} We sample $z_{la}$
from their prior distributions and $y$ from the distribution of data,
i.e. 
\begin{itemize}
\item Sample a batch of latent vectors $z_{la}\sim p_{prior}(z)$ and attribute
vectors $y\sim p_{data}(y)$. 
\item Update the SL-GAN by minimizing the Decoder $\mathcal{L}_{dec}$ and
Discriminator with $\mathcal{L}_{D_{SL-GAN}}$ iteratively; 
\end{itemize}
Once the network is trained, we can solve the task of generation and
modification; particularly, 
\begin{itemize}
\item Generating new facial images with any attributes. This can be achieved
by sampling $z_{la}$ from $p_{prior}\left(z\right)$ and setting
$y$ to any desired attributes. We can then get from the generator
the image $x'\sim G\left(z_{la},y\right)$ . 
\item Modifying the existing images with any attributes. Given an image
$x$ and the desired attribute $y$, we can sample $z_{la}\sim p\left(z\mid x\right)$.
Then the modified image can be generated by $x'\sim p_{dec}\left(x\mid z_{la},y\right)$. 
\end{itemize}

\section{Experiments}

\begin{figure*}
\centering{}%
\begin{tabular}{cc}
\includegraphics[scale=0.34]{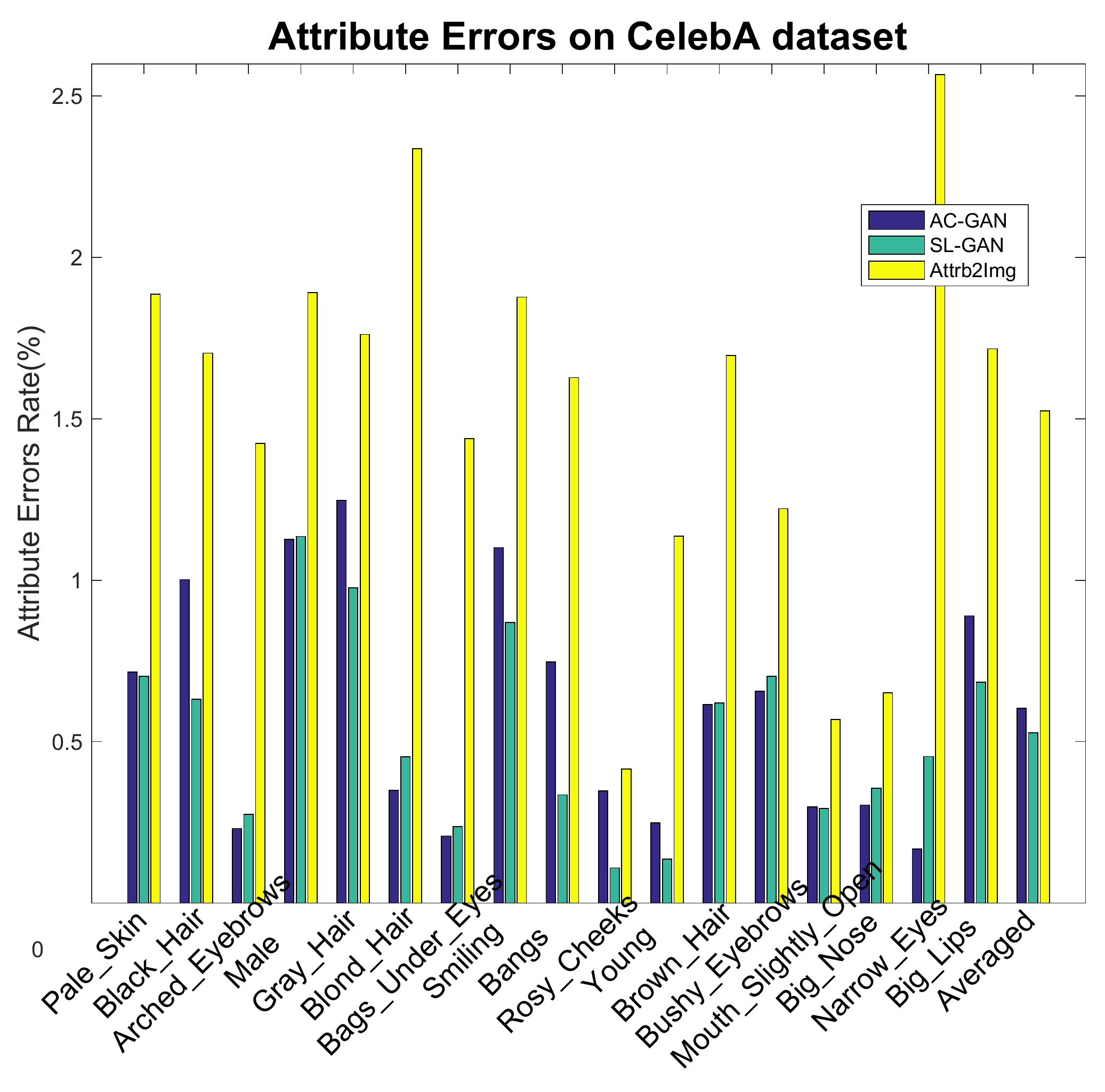}  & \includegraphics[scale=0.36]{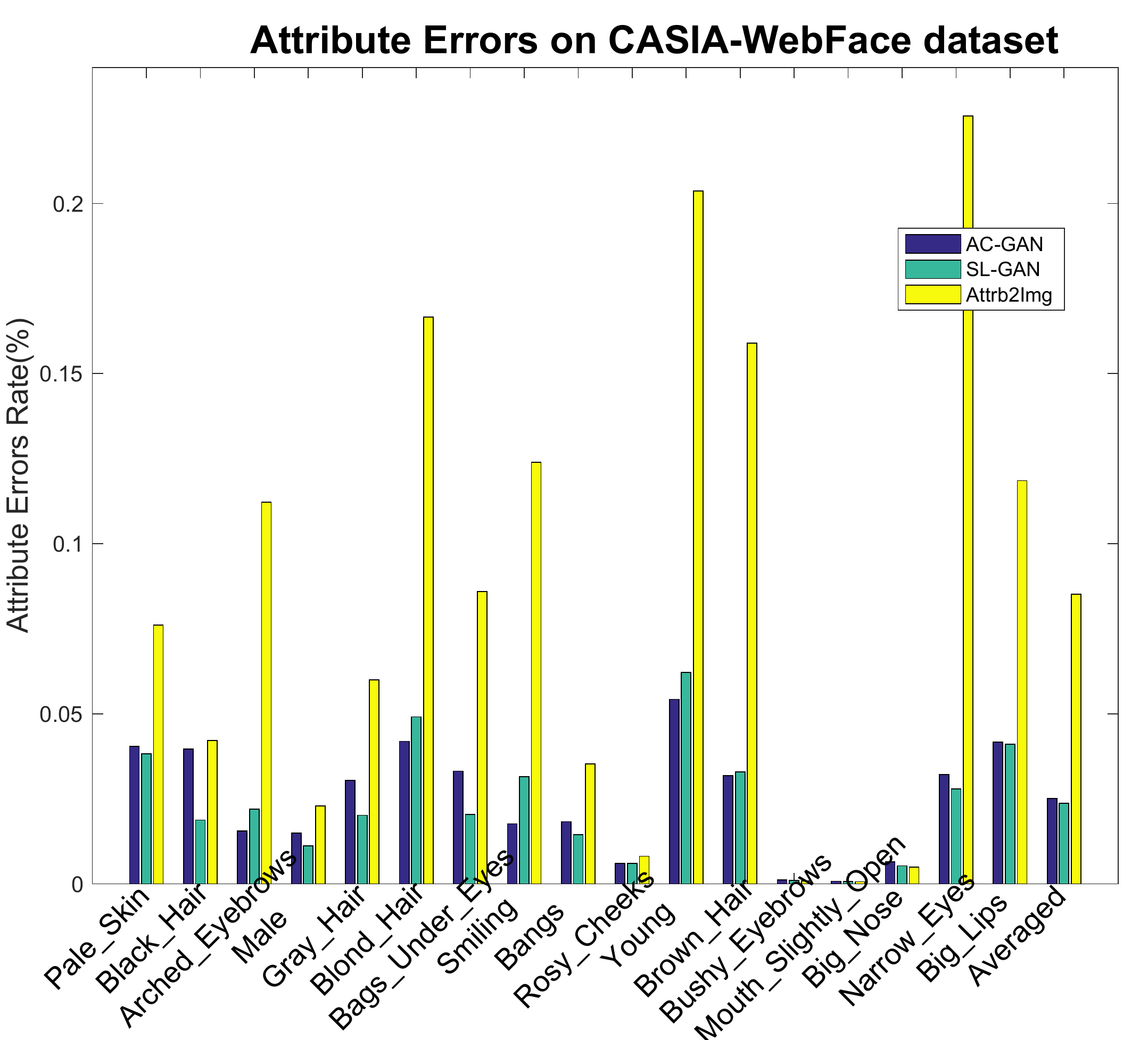}\tabularnewline
\end{tabular}\caption{\label{fig:Attribute-CelebA}Attribute Errors of user-defined attributes
on CelebA and CASIA- dataset. The lower values, the better results.
The attribute names are listed in the X-axis.}
\end{figure*}

\subsection{Experimental settings}

\begin{figure}
\centering{}\includegraphics[scale=0.23]{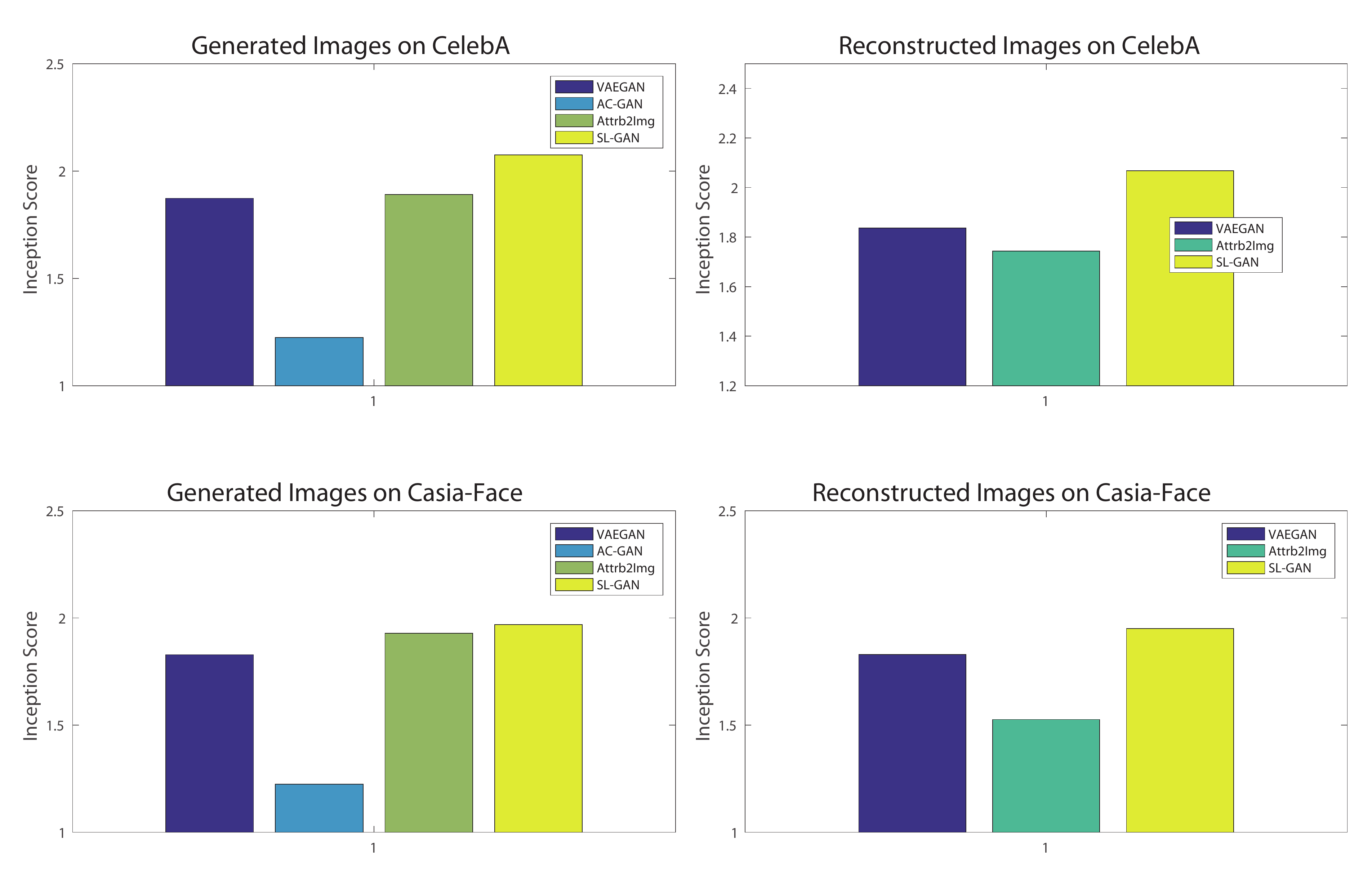}\caption{\label{fig:Inception-Scores-on}Inception Scores on two datasets.
The higher values, the better results.}
\end{figure}

\noindent \textbf{Dataset}. We conduct the experiments on two datasets.
The \emph{CelebA} dataset \cite{liu2015deep} contains approximately
$200k$ images of $10k$ identities. Each image is annotated with
$5$ landmarks (two eyes, the nose tips, the mouth corners) and binary
labels of 40 attributes. Since the distribution of the binary annotations
of attributes is very unbalanced, we select 17 attributes with relatively
balanced annotations for our experiments. We use the standard split
for our model: first 160k images are used for training, 20k images
for validation and remaining 20k for test. \emph{CASIA-WebFace} dataset
is currently the largest publicly released dataset for face verification
and identification tasks. It contains $10,575$ celebrities and $494,414$
face images which are crawled from the web. Each person has $46.8$
images on average. The CASIA-WebFace \cite{yi2014learning} doesnot
have any facial attribute labels. We use the 17 attributes of CelebA
dataset \cite{liu2015deep} to train the facial attribute classifier,
\emph{i.e.}, MOON model \cite{moon_attrb}. The trained model can
be utilized to predict the facial attributes on the CAISIA-WebFace
dataset. The predicted results are used as the facial attribute annotation.
{\em We will release these annotations, trained models and code
upon acceptance.}

\vspace{0.07in}
\noindent \textbf{Evaluation. }We employ different evaluation metrics. (1) For
the generation task, we utilize inception score and attribute errors.
\emph{Inception score} \cite{tim2016nips} measures whether varied
images are generated and whether the generated images contains meaningful
objects. Also, inspired by the attribute similarity in \cite{yan2016eccv},
we propose the \emph{attribute error}. Specifically, we uses the MOON
attribute \cite{moon_attrb} models to predict the user-defined attributes
of generated images and the predicted attributes are compared against
the specified attributes by mean square error. (2) For the modification
task, we employ the user-study for the evaluation.

\vspace{0.07in}
\noindent \textbf{Implementation details.} Training converges in 9-11 hours
on CelebA dataset on GeForce GTX 1080; our model needs around 2GB
GPU memory. The input size of images is $64\times64$. The methods
to which we compare have code available on the web, which we use to
directly compare to our results. 

\noindent 
\begin{figure}
\begin{centering}
\includegraphics[scale=0.3]{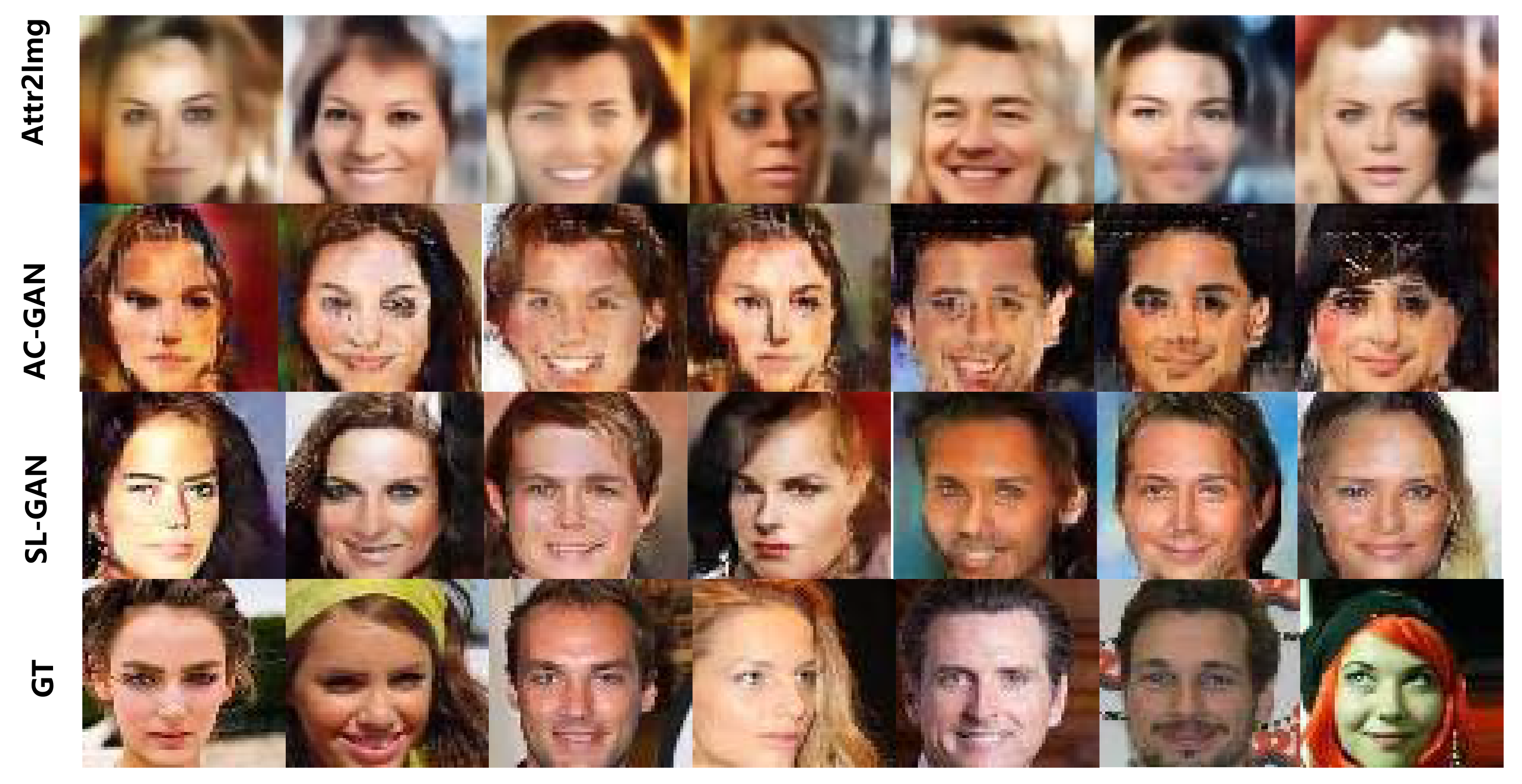} 
\par\end{centering}
\caption{\label{fig:Qualitative-results-of}Qualitative results of the generation
task. }
\end{figure}

\subsection{Generation by user-defined attributes}

\noindent \textbf{Competitors.} We compare various open-source methods
on this task, including VAE-GAN \cite{vaegan}, AC-GAN \cite{auxgan},
and Attrb2img \cite{yan2016eccv}. Attrb2img is an enhanced version
of CVAE. All the methods are trained with the same settings and have
the same number dimensions for attribute representation. For a fair
comparison, each method only trains one model on all 17 user-defined
attributes.

\noindent \textbf{Attribute errors} are compared on CelebA and CASIA-WebFace
dataset and illustrated in Fig. \ref{fig:Attribute-CelebA}. Since
VAEGAN doesnot model the attributes, our model can be only compared
against AC-GAN and Attrib2img methods. On most of the 17 attributes,
our method has lower attribute errors than the other methods. This
indicates the efficacy of our SL-GAN on learning user-defined attributes.
In contrast, the relatively higher attribute errors of Attrib2img
were largely due to the fact that Attrib2img is based on CVAE and
user-defined attributes are only given as the model input, rather
than explicitly learned as in our SL-GAN and AC-GAN. The advantages
of our SL-GAN results over AC-GAN are in part due to our generator
model with feature-wise error encoded by $\mathcal{L}_{recon}^{D_{l}}$
in Eq (\ref{eq:g_sl-gan}).

\noindent \textbf{Inception scores }are also compared on two datasets
and shown in Fig. \ref{fig:Inception-Scores-on}. We compare the inception
scores on both generated and reconstructed image settings. The differences
between generated and reconstructed images lies in how to obtain the
attribute vectors: the attribute vectors of reconstructed images is
computed by the encoder network, while such vectors of generated images
are either sampled from Gaussian prior (for $z_{la}$) or pre-defined
(for $y$).

As an objective evaluation metric, inception score, first proposed
in \cite{tim2016nips}, was found to correlate well with the human
evaluation of visual quality of samples. Thus higher inception scores
can reflect the relatively better visual quality. (1) On generated
image setting, we use the same type of attribute vector to generate
the facial images for VAEGAN, AC-GAN, and Attrib2img. On both dataset,
our SL-GAN has higher inception scores than all the other methods,
still thanks to our training algorithm for more efficiently and explicitly
learning user-defined and latent attribute in SL-FAS. These results
indicate that our generated images in general have better visual quality
than those generated by the other methods. We note that AC-GAN has
relatively lower inception scores since it does not model the feature-wise
error as our SL-GAN. (2) On reconstructed image setting, AC-GAN is
not compared since it does not have the encoder-decoder structure
to reconstruct the input image. On both dataset, our SL-GAN again
outperforms the other baselines, which suggests that the visual quality
of our reconstructed images is better than that from the other competitors.

\begin{figure*}
\centering{}\includegraphics[scale=0.98]{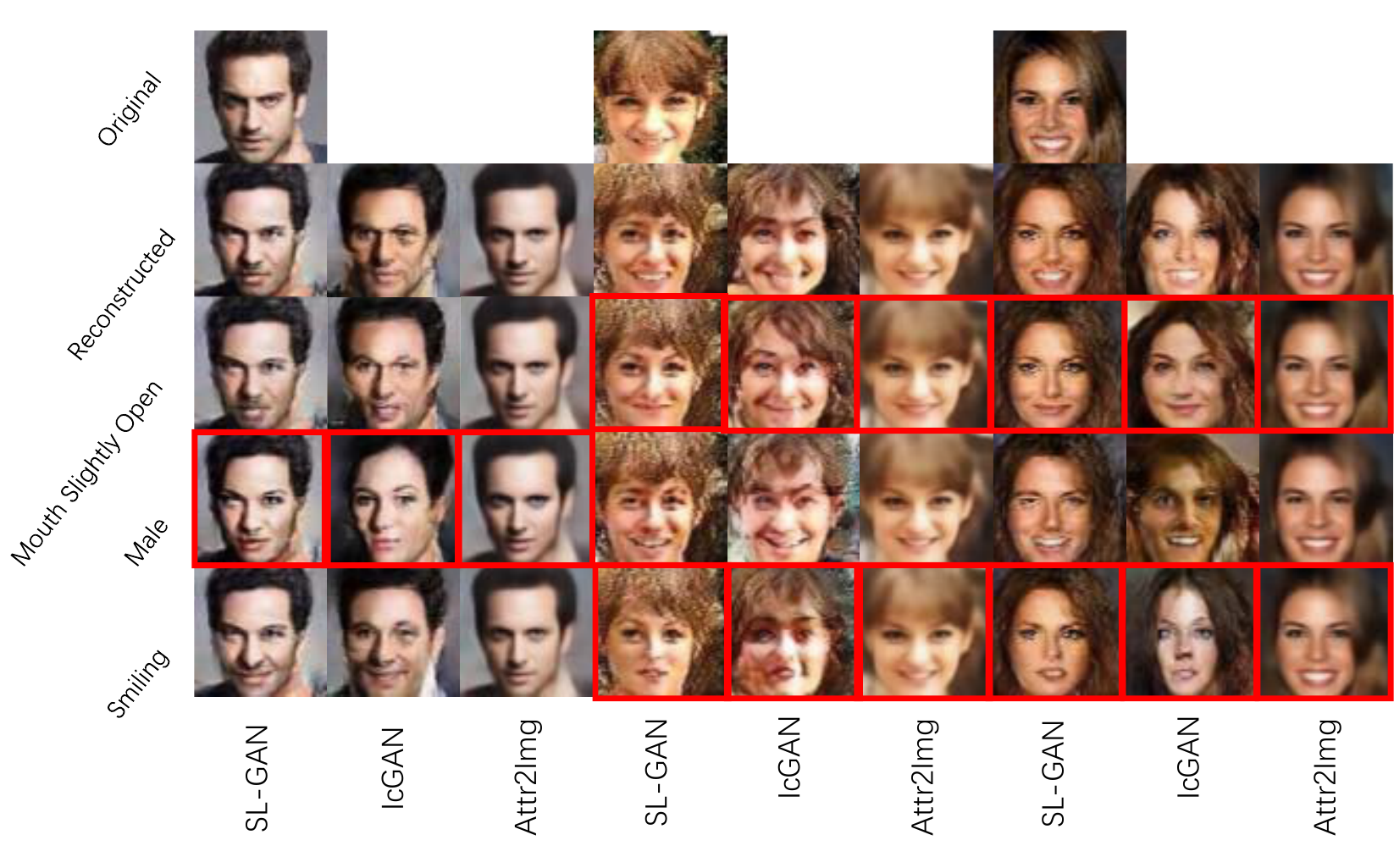}\caption{\label{fig:Qualitative-results-of-modification}Qualitative results
of comparing different modification methods. The red box indicates
that the attribute modified is inverse to the attribute of each row. }
\end{figure*}

\begin{table}
\begin{centering}
\begin{tabular}{|c|c|c|c|c|}
\hline 
Metric  & Saliency  & Quality  & Similarity  & Guess\tabularnewline
\hline 
\hline 
Attrb2img  & 3.02  & 4.01  & 4.43  & $30.0\%$\tabularnewline
\hline 
icGAN  & 4.10  & 3.83  & 4.30  & $65.4\%$\tabularnewline
\hline 
SL-GAN  & 4.37  & 4.20  & 4.45  & $75.0\%$\tabularnewline
\hline 
\end{tabular}
\par\end{centering}
\caption{\label{tab:The-user-study-of}The user-study of modification of user-defined
attributes. The ``Guess'' results are reported as the accuracy of
guessing.}
\end{table}

\noindent \textbf{Qualitative results}. Figure \ref{fig:Qualitative-results-of}
gives some qualitative examples of the generated images by VAEGAN,
AC-GAN, Attrib2img and SL-GAN as well as the ground-truth images.
For all the methods, the same attribute vectors are used for all methods
to generate the images. The generated images are compared against
the ground-truth image which is annotated by the corresponding attribute
vector. As we can see, samples generated by Attrib2img has successfully
generated the images with rich and clear details of human faces and
yet very blurred hair styles\footnote{This point is also consistency with the example figures given \cite{yan2016eccv}
which has blurred hair style details.}. In contrast, the image generated by AC-GAN can model the details
of both human faces and hair styles. However, it has lower visual
quality than our SL-GAN, which generates a more consistency and nature
style of human faces and hair styles.

\begin{figure*}
\begin{centering}
\includegraphics[scale=0.34]{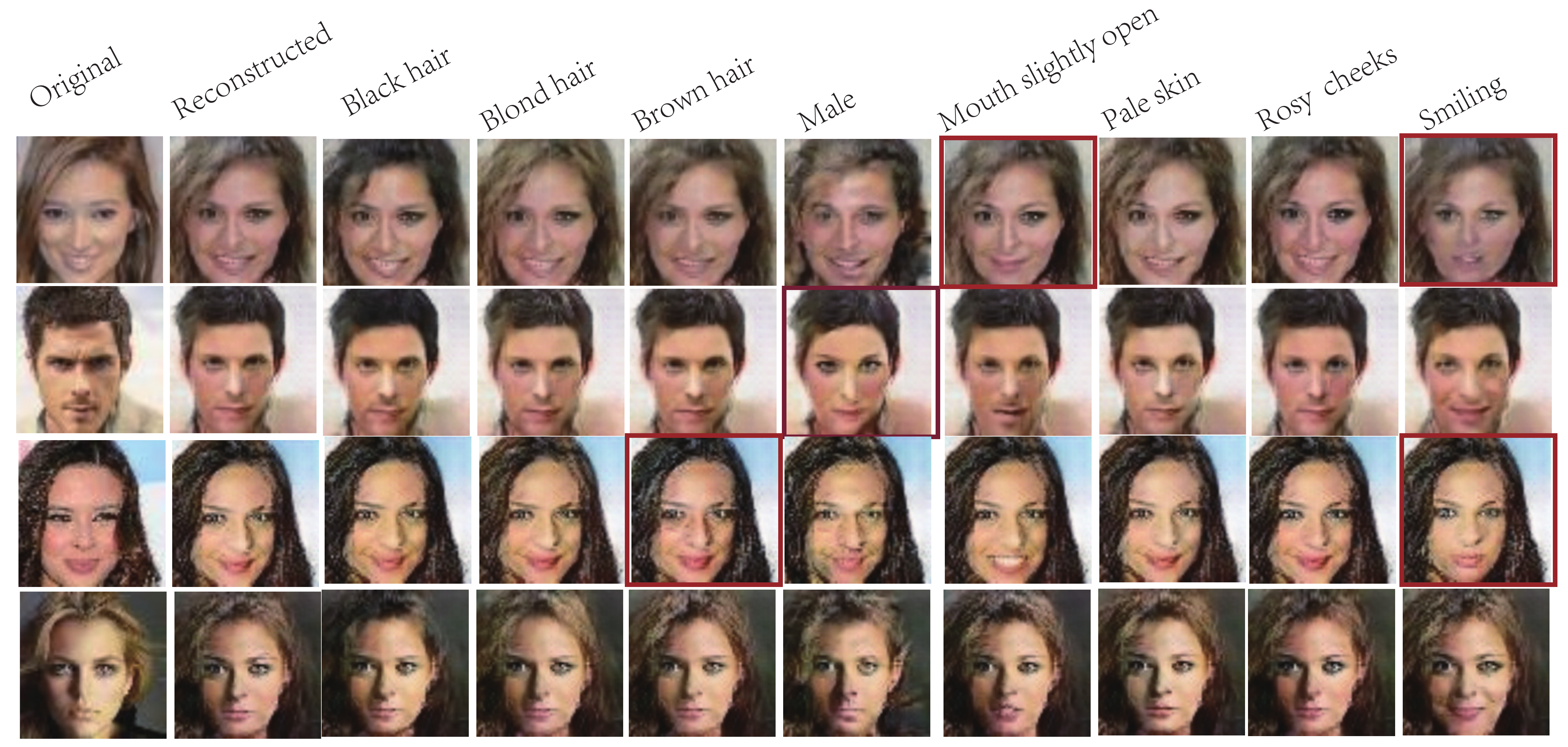} 
\par\end{centering}
\caption{\label{fig:Results-of-celebA}Results of modifying attributes on CelebA
dataset. Each column indicates modifying by adding one type of attribute
to image, while the red box means the image is modified to not have
that attribute. }
\vspace{-0.1in}
 
\end{figure*}

\subsection{Modification by user-defined attributes}

% \begin{figure}
% \centering{}{\small{}\includegraphics[scale=0.25]{fig/visualization/casia_vis}\caption{\label{fig:Results-of-casia}Results of modifying attributes on CASIA-WebFace
% dataset. Each column indicates modifying by adding one type of attribute
% to image, while the red box means the image is modified to not have
% that attribute. }
% } 
% \end{figure}

\noindent \textbf{Competitors}. We compare various open source methods
on this task, including attrib2img \cite{yan2016eccv} and icGAN \cite{Guim2016nipsworkshop}.
Note that attrb2img can not directly modify the attributes of images.
Instead, we take it as an ``attribute-conditioned image progression''
which interpolates the attributes by gradually changing the values
along attribute dimension. We still use the same settings for all
the experiments.

\noindent \textbf{User-study experiments.} We design a user study
experiment to compare attrib2img with our SL-GAN. Specifically, ten
students who are kept unknown from our projects are invited for our
user study. Given one image, we employ attrib2img and SL-GAN to modify
the same attribute and obtain two images; totally $100$ images are
sampled and the results are generated. We ask the participants to
compare each generated image with the original given image; and they
should rate their judgement on a five-point scale from the less to
the most likely by the following metrics: (1) \emph{Saliency}: the
salient degree of the attributes that has been modified in the image.
(2) \emph{Quality}: the overall quality of generated image; (3) \emph{Identity}:
Whether the generated image and the original image are the same person.
(4) \emph{Guess}: we also introduce a guessing game as the fourth
metrics. Given one modified image and the original image, we ask the
participants to guess which attributes have been modified from four
candidate choices; among these candidate choices, only one is the
correct answer (chance = $0.25$).

\noindent \textbf{Results.} The user-study results are compared in
Tab. \ref{tab:The-user-study-of}. Our results of SL-GAN outperforms
those of Attrb2img on all the metrics. These results suggest that
our SL-GAN can not only saliently modify the image attribute given,
but also guarantee the overall quality of generated image and keep
the same identity as the original input image. To better understand
the difference of different methods, we qualitatively compare these
three methods in Fig. \ref{fig:Qualitative-results-of-modification}.
As observed from the figure, our results has better overall visual
quality and more salient modified attributes.

\noindent \textbf{Qualitative results of our modification examples.
}We further show some qualitative visualization of modifying user-defined
attribute as illustrated in Fig. \ref{fig:Results-of-celebA}. Please
refer to supplementary material for larger size figures. All the attributes
are trained by only one SL-GAN model. In fact, our model can not only
change the very detailed local attribute such as ``rosy cheeks'',
``arched eyebrows'', ``Bags under eyes'', but also modify the
global attributes, such as ``Male'', ``Pale skin'', and ``Smiling''.
Furthermore, our methods are also able to change the hair styles and
hair color; and such details of hairs are usually not captured by
Attrb2img.

\section{Conclusion }

In this paper, we introduce a semi-latent facial attribute space to
jointly learn the user-defined and latent attributes from facial images.
To learn such a space, we propose a unified framework\textendash{}
SL-GAN which for the first time can both learn to generate and modify
facial image attributes. Our model is compared against the state-of-the-art
methods and achieves better performance.

{\small{}{} \bibliographystyle{ieee}
\bibliography{egbib}
 }{\small \par}
\end{document}